\documentclass[letterpaper, 10 pt, conference]{ieeeconf}  

\IEEEoverridecommandlockouts                              

\makeatletter
\let\NAT@parse\undefined
\makeatother

\usepackage{xcolor}
\usepackage{authblk}
\usepackage{graphicx}
\usepackage{amsmath}
\usepackage{subcaption}
\usepackage{booktabs}
\usepackage{comment}
\usepackage{amsfonts}
\usepackage{bbm}
\usepackage{hyperref}
\hypersetup{
    colorlinks=true,
    linkcolor=red,
    citecolor=green
}

\title{\LARGE \bf
Deep Learning a Grasp Function\\for Grasping under Gripper Pose Uncertainty
}

\author{Edward Johns, Stefan Leutenegger and Andrew J. Davison}
\affil{Dyson Robotics Lab, Imperial College London}

\begin{document}

\maketitle
\thispagestyle{empty}
\pagestyle{empty}

\begin{abstract}

This paper presents a new method for parallel-jaw grasping of isolated objects from depth images, under large gripper pose uncertainty. Whilst most approaches aim to predict the single best grasp pose from an image, our method first predicts a score for every possible grasp pose, which we denote the \emph{grasp function}. With this, it is possible to achieve grasping robust to the gripper's pose uncertainty, by smoothing the grasp function with the pose uncertainty function. Therefore, if the single best pose is adjacent to a region of poor grasp quality, that pose will no longer be chosen, and instead a pose will be chosen which is surrounded by a region of high grasp quality. To learn this function, we train a Convolutional Neural Network which takes as input a single depth image of an object, and outputs a score for each grasp pose across the image. Training data for this is generated by use of physics simulation and depth image simulation with 3D object meshes, to enable acquisition of sufficient data without requiring exhaustive real-world experiments. We evaluate with both synthetic and real experiments, and show that the learned grasp score is more robust to gripper pose uncertainty than when this uncertainty is not accounted for.

\end{abstract}

\section{INTRODUCTION}

Robot grasping is far from a solved problem. One challenge still being addressed is that of computing a suitable grasp pose, given image observations of an object. However, issuing commands to align a robot gripper with that precise pose is highly challenging in practice, due to the uncertainty in gripper pose which can arise from noisy measurements from joint encoders, deformation of kinematic links, and inaccurate calibration between the camera and the robot.

\begin{figure}[h!]
\centering
\begin{subfigure}[t]{\linewidth}
\centering
 \includegraphics[width=0.7\linewidth]{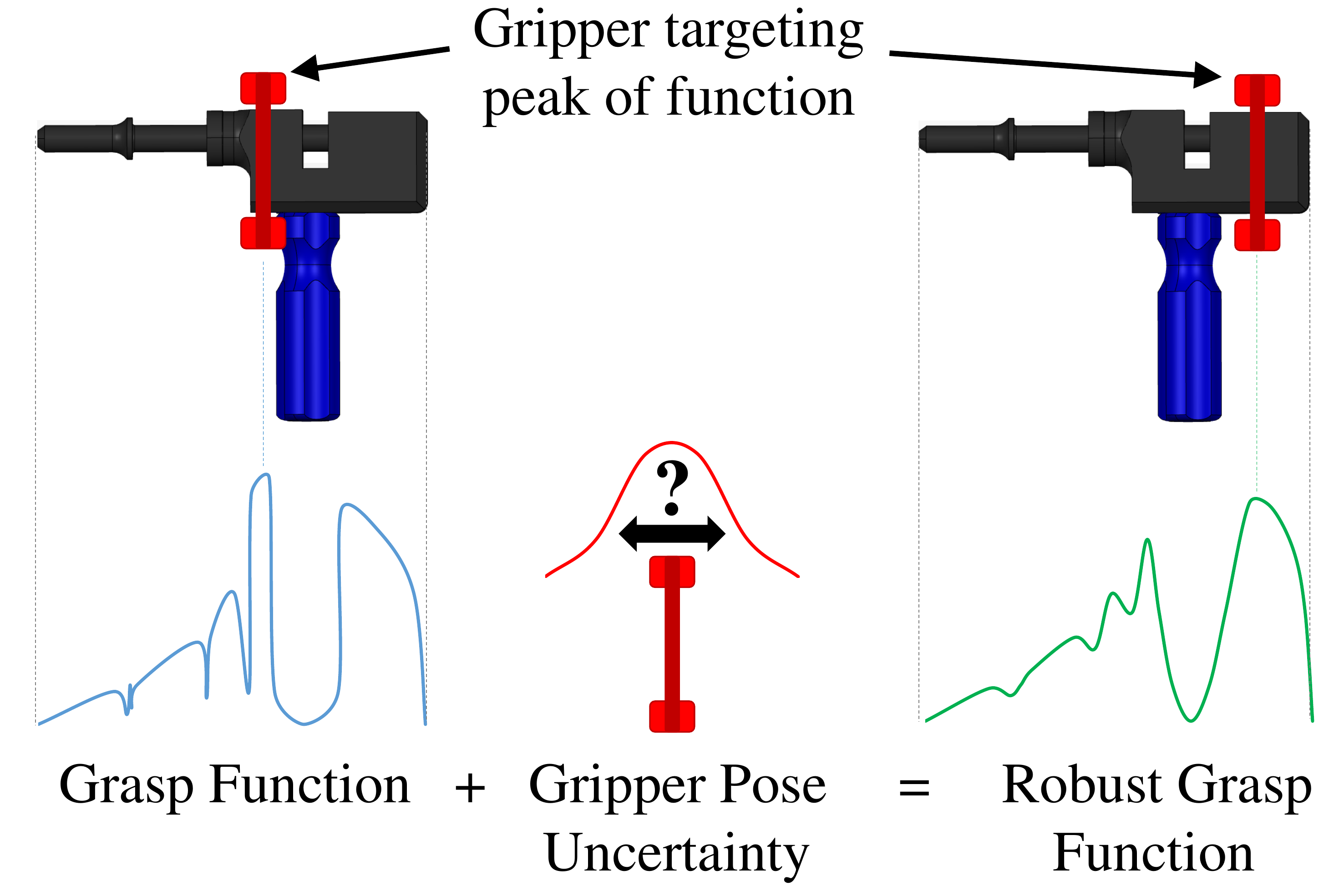}
 \caption{On the left, the gripper targets the pose corresponding to the maximum of the grasp function. On the right, the grasp function is smoothed by convolving it with the uncertainty function. By taking the maximum of this robust grasp function, it is much less likely that the robot will end up grasping in a region of poor grasp quality.}
 \label{fig:summary_a}
 \end{subfigure}
 \par\bigskip
 \begin{subfigure}[t]{\linewidth}
 \centering
  \includegraphics[width=0.3\linewidth]{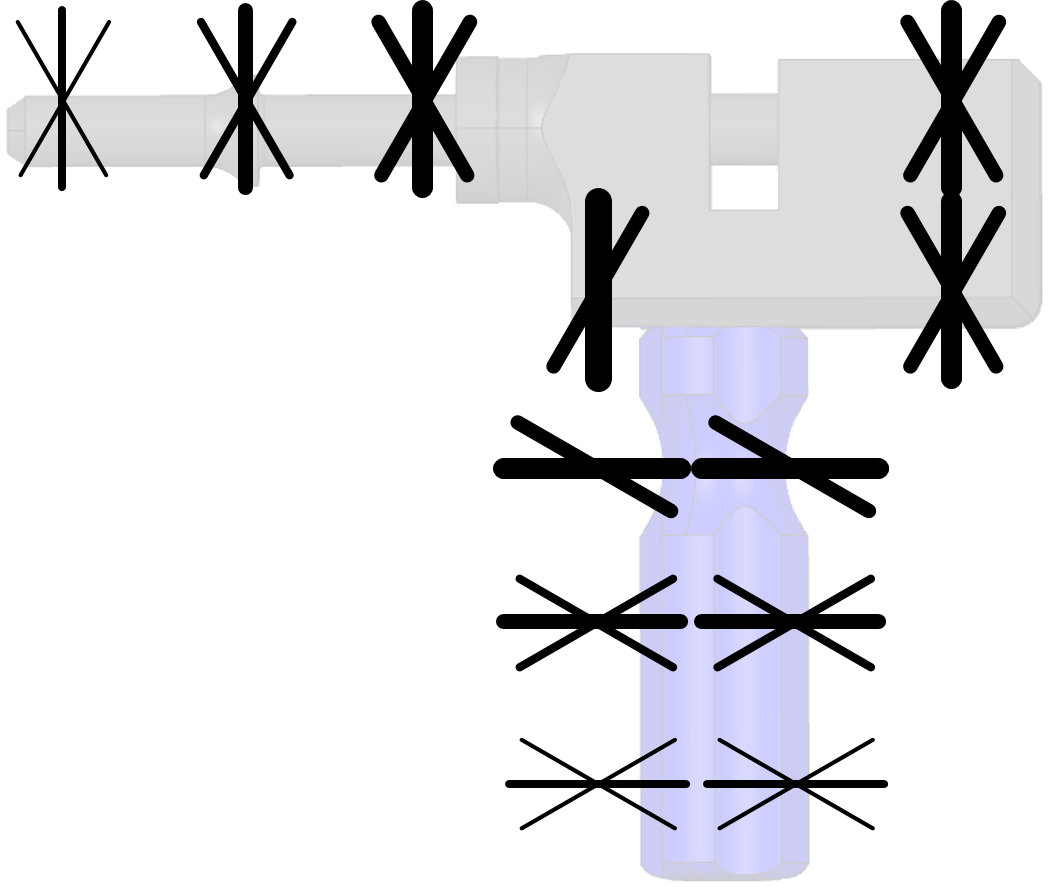}
 \caption{Visualisation of the grasp function. Each line represents a gripper pose, where the fingers start at the ends of the line and move towards the centre of the line. The thickness of each line indicates the grasp score for that pose.}
 \label{fig:summary_b}
 \end{subfigure}
 \caption{Introducing the role of the grasp function. In (a), the grasp function is a function of horizontal displacement for ease of visualisation, whereas the grasp function we learn fully explores 2D space, as in (b).}
\end{figure}

Consider attempting a grasp on the object in Figure \ref{fig:summary_a}. Given perfect control of a robot arm, the maximum grasp quality across the object could be targeted (peak of blue function). However, if the gripper misses its target, then in this case it will achieve a very poor grasp; to the left, there are unstable regions, and to the right, there is a part of the object which would block the grasp and cause a collision.

To solve this, rather than predicting a single grasp, we propose to learn a \emph{grasp function}, which computes a grasp quality score over all possible grasp poses, given a certain level of discretisation. Here lies our key novelty, and this allows for the gripper's pose uncertainty to then be marginalised out, by smoothing the grasp function with a function representing this uncertainty, to yield a \emph{robust grasp function}. In this way, the final grasp pose will target an area of the object which no longer lies directly next to areas of poor grasp quality, as shown in the right of Figure \ref{fig:summary_a}.

To generate the grasp function, we train a Convolutional Neural Network (CNN) to predict the grasp score for every pose of a parallel-jaw gripper, with respect to an observed depth image of an object. This is visualised in Figure \ref{fig:summary_b}, whereby each line indicates a gripper pose. To satisfy the need of CNNs for large volumes of training data, we generate training pairs in simulation, by rendering synthetic depth images of 3D meshes, and use a physics simulator to predict the quality of grasps over the range of gripper poses. Whilst this uncertainty could be incorporated directly into the simulation, learning the grasp function enables different arms to operate with just one set of simulations, and also allows for incorporation of dynamic uncertainty if particular arm configurations are known to have different uncertainties. Furthermore, the grasp function could have broader use in the context of grasp planning, when obstacles or arm kinematics may prevent the achievement of some poses.

\section{RELATED WORK}

In recent years, deep learning \cite{r4} and its computer vision counterpart CNNs \cite{r5} have revolutionised the fields of object recognition \cite{r27}, object segmentation \cite{r28}, and local feature learning \cite{r23}. Consequently, these methods have also shown to be successful in robotics applications. Robot localisation is moving away from using hand-engineered features \cite{r29} and towards deep learning features \cite{r31},  active recognition has achieved state-of-the-art performance by deep learning camera control \cite{r2}, deep reinforcement learning is enabling end-to-end training for robot arm control \cite{r25}, and even autonomous driving has been tackled by similar learning-based approaches \cite{r24}.

For determining object grasp poses from images, use of hand-engineered features still performs well in complex cases such as dense clutter \cite{r30} or multi-fingered grasping \cite{r17}. However, in simpler cases, grasp pose detection via CNNs has achieved state-of-the-art solutions. One approach to this has been to train on manually-labelled datasets, where human labellers have determined the location of a suitable grasp point on an image of an object. The Cornell grasping dataset \cite{r7} for example, consists of RGB and depth images with a parallel-jaw gripper pose defined in image coordinates. In \cite{r6}, a CNN is trained on this dataset by combining both RGB and depth data into a single network, and predicting the probability that a particular pose will be graspable, by passing the corresponding image patch through the network. This method was speeded up in \cite{r8} by passing the entire image through the network rather than individual patches, eliminating the time-consuming need to process multiple patches for each frame.

One challenge with deep learning is the need for a very large volume of training data, and the use of manually-labelled images is therefore not suitable for larger-scale training. One alternative approach, which we also adopt, has been to generate training data in simulation, and attempt to minimise the gap between synthetic data and real data. A popular example is the GraspIt! simulator \cite{r3}, which processes 3D meshes of objects and computes the stability of a grasp based upon the grasp wrench space. Whilst these methods do not incorporate dynamic effects which are typically involved in real grasping, prediction of static grasps can be achieved to a high accuracy by close analysis of the object shape. In \cite{r9}, this simulation was used to predict the suitability of a RGBD patch for finger locations in multi-fingered grasping, together with the suitability of each type of hand configuration. The work of \cite{r10} used a similar static grasp metric and considered uncertainty in gripper pose, object pose, and frictional coefficients.  In \cite{r32}, grasping in clutter was achieved by using a static stability heuristic, based on a partial reconstruction of the objects.

Static metrics for generating training data have their limitations though, due to the ignorance of motion as the object is lifted from the surface. In \cite{r11,r15,r16}, it has been shown that dynamic physics simulations offer a more accurate prediction of grasp quality than the standard static metrics. Furthermore, \cite{r11} illustrated how good-quality grasps predicted by physics simulations are highly correlated with those predicted by human labellers.

One final approach to generating training data for deep learning is to do so with real-world experiments on a real robot. It was shown in \cite{r12} that a reinforcement learning approach can achieve effective results by testing grasps on a real platform, although training time was several weeks and thus the scalability is of great limitation. \cite{r22} then scaled this up from weeks to months, and also used multiple robots in parallel to learn a form of visual servoing, to predict when a moving gripper is in a suitable pose to grasp the object currently between the gripper's jaws. We avoid these approaches due to their lack of scalability and flexibility to challenges more complex than simple parallel-jaw grasping, and investigate how well a simulation can model real-world behaviour.

For parallel-jaw grasping as in our work, all the prior solutions fall into one of three categories: regressing the optimum grasp from an entire image, regressing the optimum grasp from a patch, or assigning a grasp quality score to patch. In this paper, we present the first work, to our knowledge, which effectively follows the third approach, but does so directly from a single image without processing individual patches, and hence achieving real-time grasping. This prediction of grasp quality scores over the entire image then allows us to incorporate gripper pose uncertainty during online operation, by smoothing the grasp quality score distribution with this uncertainty.

\section{DEFINING A TARGET POSE}

\begin{figure}[h]
    \centering
    \includegraphics[width=0.8\linewidth]{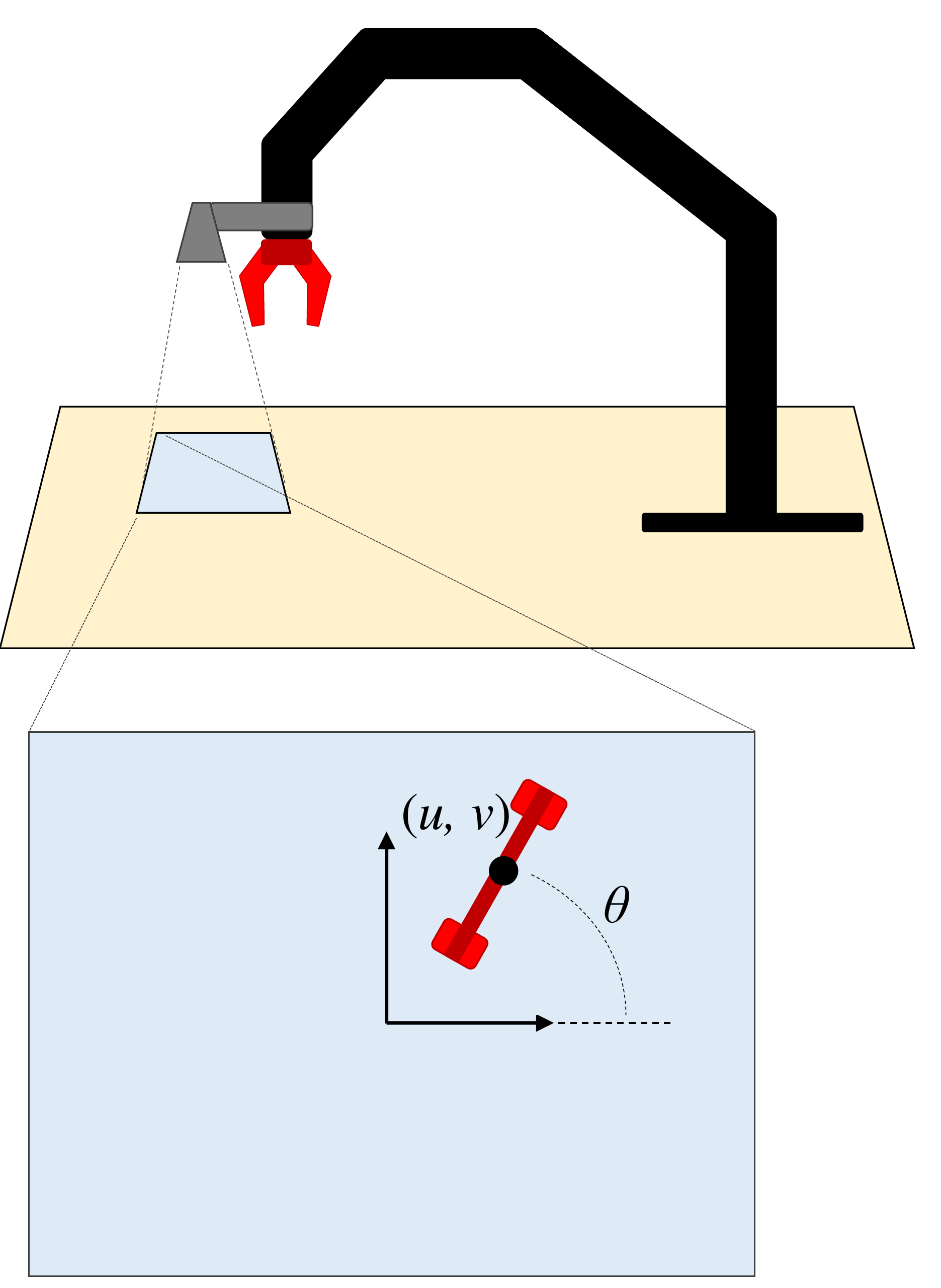}
    \caption{The gripper's target pose is defined in image coordinates, with a depth camera mounted on the robot's wrist. This pose is then transformed into the robot's frame via a calibration between the camera and the gripper.}
    \label{fig:gripper_pose}
\end{figure}

The task is to grasp a single isolated object with a parallel-jaw gripper, by observing a depth image of the object from a single view and computing an optimum pose for the gripper to be sent to, as depicted in Figure \ref{fig:gripper_pose}. We follow the trend in this setup \cite{r6, r8, r12, r22} by constraining the gripper to a perpendicular orientation with respect to a flat table surface upon which the object is resting. Grasps are then executed at a constant height such that the gripper's tips are 1mm above the surface at the lowest point in the grasp. The remaining parameters to be learned which define the grasp pose, are therefore the translational position on the surface, and the rotation of the gripper with respect to the surface's normal.

Given that the target gripper pose is computed through image observations, we first define these free parameters of the target pose in terms of image coordinates. In our experiments, we rigidly mount a depth camera onto the wrist of the robot arm, although this could instead be achieved by additional apparatus as in \cite{r8, r12}. The camera is positioned at a fixed height from the surface, and with viewing direction parallel to the surface's normal. By calibrating this camera with respect to the robot, a target pose in the robot frame can therefore be computed and executed. In these image coordinates, we denote the target pose as $p = \{u, v, \theta\}$, where $u$ and $v$ are the horizontal and vertical translations of the gripper's centroid relative to the centre of the image, and $\theta$ is the rotation of the gripper about the image's $z$-axis, as shown in Figure \ref{fig:gripper_pose}. Given the static transformation between the camera frame and the gripper frame, we can transform a target pose in image coordinates to a target pose in the robot's frame by a sequence of transformations:

\begin{equation}
\textrm{T}_{RP} \; = \; \textrm{T}_{RG} \; \textrm{T}_{GC} \; \textrm{T}_{CI} \;\; p,
\end{equation}

where $\textrm{T}_{RP}$ is the target gripper pose in the robot frame, $\textrm{T}_{RG}$ is the transformation between the robot frame and the starting gripper frame (when the image was captured), $\textrm{T}_{GC}$ is the calibrated transformation between the gripper frame and the camera frame, and $\textrm{T}_{CI}$ is the transformation between 2D image coordinates and the 3D camera frame.

To make training and inference tractable, the space of gripper poses defined by $p$ is now discretised into space $Q$, such that each element $q \in Q$ represents a unique gripper pose (one line in Figure \ref{fig:4c}). Choice in the granularity of this discretisation is a compromise between precision in target pose prediction, and tractability of training. We chose a discretisation of 1cm in translation on the table surface, corresponding to a 14 pixels in the $640\times 480$ images, and $30 ^{\circ}$ in rotation. The task now becomes to predict, from an observed depth image, a single grasp quality score for each of these 8712 possible grasp poses in $Q$, to yield the overall grasp function $f(q)$.

\section{GENERATING TRAINING DATA}

Prediction of these grasp quality scores is achieved by training a CNN to take as input a single depth image, and to output a score for each grasp pose. CNNs require a huge amount of training data to fully exploit their capacity to learn complex functions, whilst avoiding overfitting. As such, real-world experiments are not scalable enough to generate the required extent of data to learn the grasp function. Instead, we generate all our training data in simulation, by rendering depth images using OpenGL's depth buffer, and using a physics engine to simulate grasps. We capitalise on the recent ModelNet dataset \cite{r1}, a large-scale collection of 3D mesh models representing a range of common objects which has been collated specifically for deep learning experimentation. The use of mesh models is particularly suitable for our application, for two reasons. First, synthetic depth images can be easily rendered from meshes, which are much more realistic than synthetic RGB images, as these often struggle to model illumination and texture with sufficient realism. Second, mesh models allow us to directly attempt grasps over the entirety of the model using physics simulation.

\subsection{Physics Simulation}

We use the Dynamic Animation and Robotics Toolkit (DART) \cite{r26} for physics simulation. DART was chosen over other static simulators such as with GraspIt! \cite{r3}, due to the importance of dynamics modelling in predicting the behaviour of real-world grasps, which typically involve movement after a grasp is executed. DART was also chosen because of its suitability for robotics applications from its hard constraints imposed on the resolving of forces, its support of mesh models for collision detection, and its intuitive use of generalised coordinates for kinematics modelling. Whilst DART is slower than physics engines more suited to computer graphics, such as Bullet, its greater accuracy in modelling precise, real-world behaviour is imperative for an application such as grasp simulation. However, a range of alternative physics simulators also exist which could be interchangeable for the physics simulation \cite{r18}.

\begin{figure}[h]
\centering
 \includegraphics[width=\linewidth]{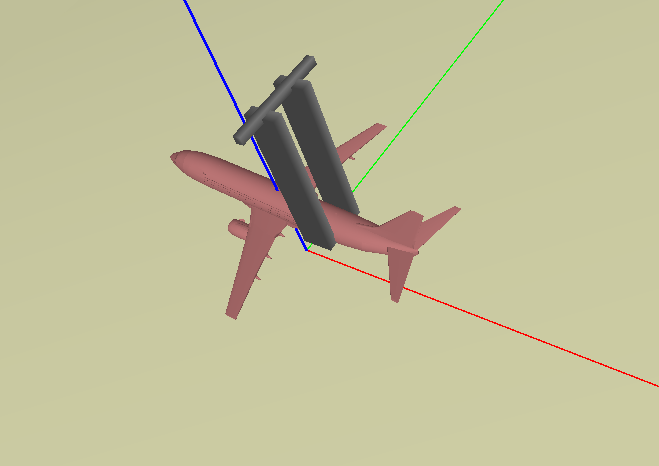}
 \caption{We use physics simulation to execute parallel-jaw grasps upon a 3D object mesh. The above shows a successful grasp, whereby the object is lifted off the surface by 20cm. Simulated grasps are attempted on the object over all gripper poses.}
 \label{fig:dart_plane}
\end{figure}

Using DART, we constructed a simple parallel-jaw gripper, consisting of a "hand" which is free to move kinematically, and two "fingers" which are controlled dynamically by torque control on a revolute joint. We then constructed a flat surface, upon which each object mesh is placed. To execute a grasp, the hand is positioned in a specified pose on the surface, after which a constant force is applied to the finger joints for a fixed time. Subsequently, the hand is raised upwards by 20 cm at a constant speed of 0.1 m/s. The dimensions of the gripper were matched to that of our Kinova MICO arm, with a fixed distance of 10cm between fingers. Figure \ref{fig:dart_plane} demonstrates the simulation world for a executing a single grasp attempt.

Due to imperfections in the physics simulator, physical values including the applied finger torque, the coefficients of friction, and the object density, were all manually tweaked to yield an acceptable level of grasping realism over a range of objects. Whilst this would be an obvious flaw if these properties were being modelled, in our case the important output is the relative grasp quality over different poses -- not the absolute magnitude of torque required.

\subsection{Image Simulation}

To simulate a depth image of the object model, images were rendered using OpenGL's depth buffer. A noise model, inspired by \cite{r19}, was then applied to simulate a true image from the Primesense Carmine 1.09 camera used in our experiments. This noise model consists of two Gaussian components: random shifting of pixels in a local region to simulate noise in the localisation of each depth measurement, and further random noise to simulate noise in the depth measurements themselves. For a ground truth depth of $z(u, v)$ at pixel location $(u, v)$, the depth after applying the model is defined as:

\begin{equation}
\hat{z}(u, v) = z(u + \mathcal{N}(0, \sigma_p^2), v + \mathcal{N}(0, \sigma_p^2)) + \mathcal{N}(0, \sigma_d^2),
\end{equation}

where $\sigma_p$ is the standard deviation of noise in pixel localisation (set to 1 pixel in our experiments), and $\sigma_d$ is the standard deviation of noise in depth estimation (set to 1.5mm in our experiments). This noisy image is then used as input for training the CNN. Figure \ref{fig:4b} illustrates the effect of applying this noise model to a synthetic depth image.

\subsection{Grasp Synthesis}

\begin{figure*}
    \centering
    \begin{subfigure}[t]{0.23\textwidth}
        \includegraphics[width=\textwidth]{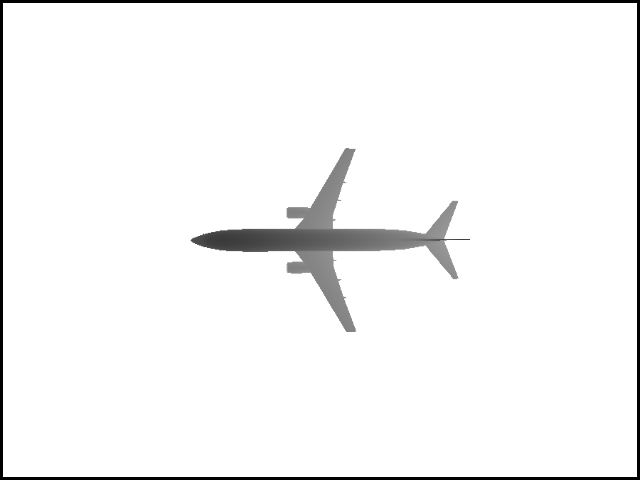}
        \caption{Simulated depth image}
    \end{subfigure}
    \hspace{2mm}
    \begin{subfigure}[t]{0.23\textwidth}
        \includegraphics[width=\textwidth]{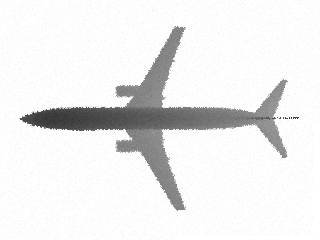}
        \caption{Simulated depth image after adding the noise model}
        \label{fig:4b}
    \end{subfigure}
    \hspace{2mm}
    \begin{subfigure}[t]{0.23\textwidth}
        \includegraphics[width=\textwidth]{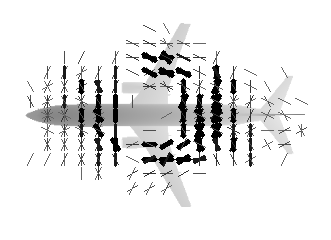}
        \caption{Visualisation of the grasp function}
        \label{fig:4c}
    \end{subfigure}
    \hspace{2mm}
    \begin{subfigure}[t]{0.23\textwidth}
        \includegraphics[width=\textwidth]{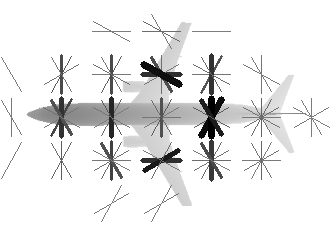}
        \caption{Visualisation of the grasp function at a coarser scale}
    \end{subfigure}
    \caption{Illustrating the grasp function during training. (a) shows the ground-truth synthetic depth image, and (b) shows a zoomed-in image after applying the noise model. (c) then visualises the grasp function overlayed on the (zoomed-in) depth image which was computed by the physics simulation. Each line represents a grasp position and direction, and the line thickness indicates the score for that pose. (d) then shows a similar visualisation to (c), but with a coarser distribution of poses, although the discretisation in (c) is the true level of granularity used in our work.}
    \label{fig:noise}
\end{figure*}

A ground truth grasp function is then calculated for every training object, by computing the grasp quality score $f(q)$ for every discrete pose $q$. Each object was placed at a position in the centre of the camera image, and grasp attempts were executed for each pose. If the object was successfully lifted fully off the surface to a height of 20cm, then that attempt was labelled with a score of 1. Each pose was assigned five grasp attempts, with each defined by a uniformly random draw from the continuous space of poses encompassing that discrete pose. The overall score for the pose was then the average over all five attempts. Adding this small random noise allows a more informative score to be assigned to the pose from a range of 0 to 1 in 0.2 intervals, rather than a less informative score of either 0 or 1. Figure \ref{fig:4c} presents a visualisation of these scores across all poses for one training object. As can be seen, higher scores are assigned to those areas of the image corresponding to stable areas for grasping.

Objects were selected from the ModelNet database \cite{r1} for training, covering a range of shapes such as \emph{airplane}, \emph{chair}, \emph{dish} and \emph{hammer}. For each model, rather than placing it upright in its standard pose, a random orientation was chosen. This is because rather than learning about object identity or semantics, pure grasping is more concerned with learning an understanding of object shape, and enriching the training images with observations from a wide range of unusual shapes is much more important than learning pose-specific grasps. Furthermore, we randomly scaled each object model to within a range of appropriate dimensions for graspable objects, with the maximum dimension constrained to being between 5cm and 20cm.

\section{LEARNING THE GRASP FUNCTION}

To learn a mapping between a depth image and the predicted grasp score $f(\hat{q})$, we train a CNN to output a score for every pose $\hat{q} \in Q$ over the image. Recent developments in deep learning now allow for networks which can learn very complex functions with high-dimensional outputs. We exploit this to enable prediction of a grasp score as a distribution over poses, rather than, for example, just predicting the single pose with the maximum score, as is typical with traditional learning-based grasping solutions. However, whilst deep networks can be trained for direct regression \cite{r20}, their performance is superior when trained for classification, and so we form the training problem as one of classifying each pose in terms of its score.

\begin{figure*}
\centering
 \includegraphics[width=\linewidth]{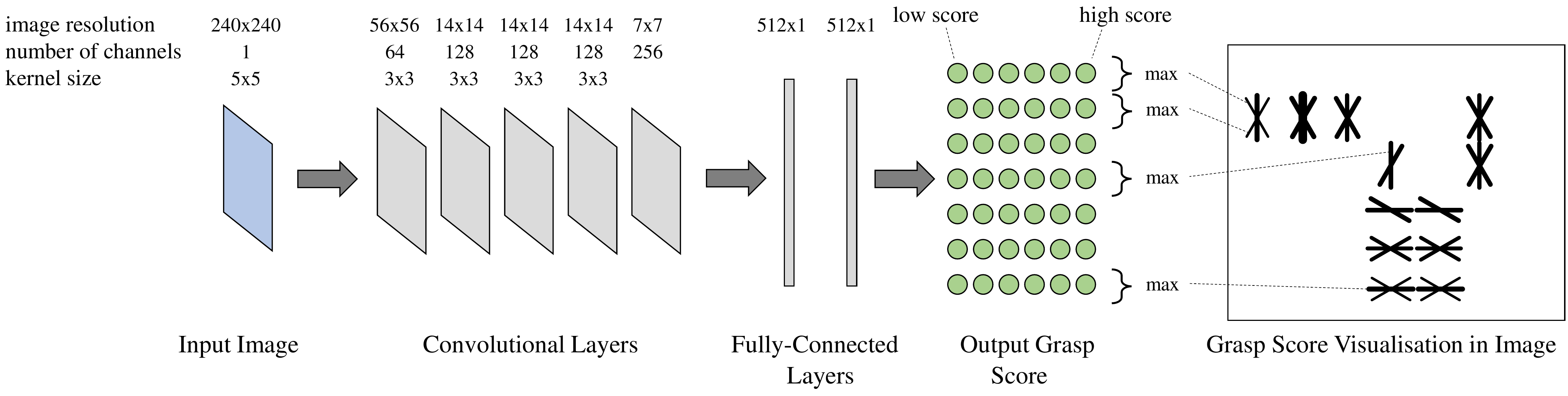}
 \caption{The structure of our CNN used for predicting the grasp function. Each input image is mapped onto a set poses, where each pose is represented by 6 nodes, one for each level of score (0, 0.2, 0.4, 0.6, 0.8, 1.0). The visualisation on the right shows the strength of each pose, by taking the maximum of these six score levels.}
 \label{fig:cnn}
\end{figure*}

We retain the discretisation of scores already existing due to attempting five grasps per pose, and train the network to predict which of the six possible scores {0, 0.2, 0.4, 0.6, 0.8, 1.0}, the pose should be assigned to. To achieve this, a network structure similar to that of AlexNet \cite{r5} is adopted, with five convolutional layers adjoined by max pooling, and two fully-connected layers. The network takes as input a depth image, and then outputs a single value for each of the six scores. Figure \ref{fig:cnn} illustrates the structure of the network, and how the output corresponds to the learned grasp function.

The network is trained by attempting to reduce the difference, across all images and all poses, between the ground truth score class $y$ ($y \in \{0, 0.2, 0.4, 0.6, 0.8, 1.0\}$), and the score class predicted by the network $\hat{y} = \text{max}(\mathbf{\hat{y}})$. This is achieved by minimising the following loss function:

\begin{equation}
L = \sum_{i}^{B}\sum_{j}^{M}\sum_{k}^{N}\delta(k, y_{ij}) \cdot \text{softmax}(\hat{\mathbf{y}}_{ij})^k.
\end{equation}

Here, $B$ is the number of training images in a mini-batch, $M$ is the number of poses output by the network ($=|Q| = 8712$), and $N$ is the number of possible scores which the pose can be classified as (= 6). $y_{ij}$ therefore represents the ground truth score value for the $i^{th}$ image and the $j^{th}$ gripper pose, and $\mathbf{\hat{y}}_{ij}$ is a length-$N$ vector corresponding to the output from the CNN for this pose. The indicator function $\delta(k, y_{ij})$ is equal to 1 iff the score $y_{ij}$ is equal to $k$, and equal to 0 otherwise. As is standard in neural network classification, the softmax function is used here rather than the L1 or L2 norm, because only the relative values of the scores are important, not their absolute values. This is defined by:

\begin{equation}
\text{softmax}(\hat{\mathbf{y}}_{ij})^k = \frac{e^{\hat{y}_{ij}^k}}{\sum_{l}^{N} e^{\hat{y}_{ij}^l}},
\end{equation}

Training was done with gradient descent on mini-batches, using the TensorFlow library \cite{r21}.

\section{GRASP EXECUTION}

\begin{figure}
    \centering
    \begin{subfigure}[t]{0.48\linewidth}
        \includegraphics[width=\textwidth]{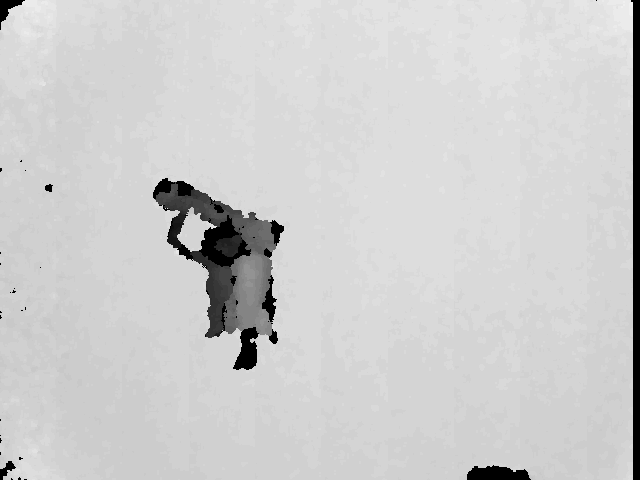}
        \caption{Observed real depth image}
        \label{fig:depth_image}
    \end{subfigure}
    ~ 
    \begin{subfigure}[t]{0.48\linewidth}
        \includegraphics[width=\textwidth]{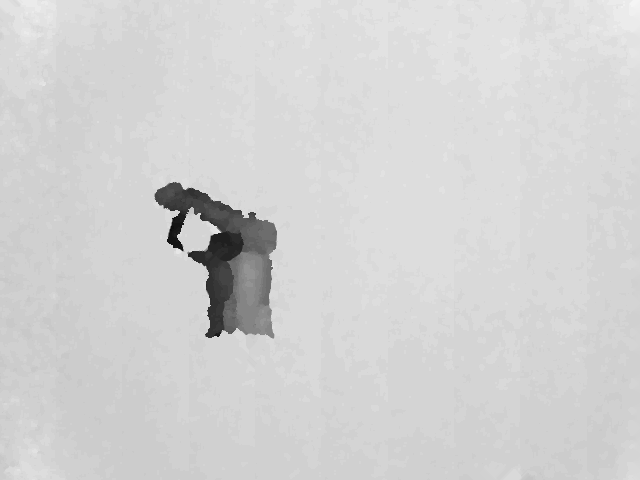}
        \caption{After replacement of ``zero'' pixels with an estimate of the surrounding depth.}
        \label{fig:noisy_depth_image}
    \end{subfigure}
    \caption{During inference, each captured depth image is preprocessed by filling in the ``zero'' pixels, which are a result of depth shadows, and bodies which do not reflect infra-red light well.}
    \label{fig:inpainting}
\end{figure}

During grasp execution on a real robot, the depth image is first preprocessed to remove pixels which have zero values, as a consequence of imperfect sensing. This is achieved by replacing those pixels with the nearest non-zero pixel, or if there are several within a given radius, the average of these. Figure \ref{fig:inpainting} illustrates this effect.

\setlength\tabcolsep{5.6pt}
\begin{table*}
\scriptsize
\centering
\begin{tabular}{c|cccc|cccc|cccc|cccc}
\toprule
    & \multicolumn{4}{c}{$\sigma_{\theta}$=10} & \multicolumn{4}{c}{20} & \multicolumn{4}{c}{30} & \multicolumn{4}{c}{40} \\
    \midrule
    Method & $\sigma_{uv}$=5 & 10 & 15 & 20 & 5 & 10 & 15 & 20 & 5 & 10 & 15 & 20 & 5 & 10 & 15 & 20 \\
\midrule
Centroid & 0.767 & 0.732 & 0.689 & 0.638 & 0.742 & 0.716 & 0.669 & 0.620 & 0.730 & 0.698 & 0.654 & 0.61 & 0.703 & 0.639 & 0.648 & 0.587 \\
Best Grasp & 0.832 & 0.789 & 0.720 & 0.612 & 0.813 & 0.772 & 0.699 & 0.587 & 0.800 & 0.767 & 0.697 & 0.555 & 0.753 & 0.740 & 0.655 & 0.548 \\
Robust Best Grasp & 0.838 & 0.813 & 0.780 & 0.756 & 0.828 & 0.801 & 0.763 & 0.697 & 0.801 & 0.779 & 0.754 & 0.688 & 0.767 & 0.768 & 0.731 & 0.656 \\
\bottomrule
\end{tabular}
\caption{Simulation results for the three methods over a range of gripper pose uncertainties. Numbers represent the percentage of objects which were successfully grasped. Uncertainties $\sigma_{\theta}$ is in degrees, whilst $\sigma_{uv}$ is in mm.}
\label{table1}
\end{table*}

The network then takes this processed depth image, and outputs a grasp function, predicting the quality of each gripper pose if a grasp was attempted thereby. Now, we assume that after attempting to send the gripper to a target pose $q$, the actual pose achieved $\hat{q}$ is defined probabilistically by a covariance matrix $\Sigma$ representing the gripper's pose uncertainty, such that $\hat{q} \sim \mathcal{N}(q, \Sigma)$. This uncertainty is defined within the $(u, v)$ plane over which the grasp function is computed. During grasp planning, the pose uncertainty is then marginalised out by convolving the grasp function with the probability density function of the achieved true gripper pose given this uncertainty. Effectively, this is performed by smoothing the grasp function in 3-dimensions $(u, v, \theta)$, with a kernel corresponding to a Gaussian distribution, whose covariance matrix is that given by the gripper pose uncertainty. Finally, trilinear interpolation is performed over pose space $(u, v, \theta)$ to achieve precision beyond the pose discretisation level, and the maximum of this final distribution is the pose to which the gripper is sent.

\section{EXPERIMENTS}

To train the network, 1000 objects were randomly selected from the ModelNet dataset of 3D object meshes \cite{r1} and processed by the simulator. This generation of training data took roughly one week to complete, at an average time per object of about 10 minutes. Whilst the object models were all placed directly beneath the gripper during physics simulation, we augmented the training data to allow for robustness to camera orientation. To achieve this, for each training image, we randomly rotated about $\theta$ and randomly shifted about $(u, v)$ to achieve 1000 new augmented images per original image. The associated grasp score for each augmented image can be easily calculated by transforming in the same way. To train the CNN, we pre-trained the weights for classification of the ModelNet40 datasest \cite{r1} before tuning the network towards learning the grasp function.

For validation of our approach, we conducted experiments in both simulation with synthetic data, and with real-world grasping on a robot platform. Whilst the real-world experiments investigate how well our synthetic training adapts to real data, the simulation experiments evaluate our method on both a much larger dataset, and over a much larger range of parameters, than could be achieved with real-world experiments.

\subsection{Simulation}

Experiments in simulation were carried out by using a similar setup as during data collection. 1000 further objects were randomly selected from the ModelNet dataset for testing. We then tested the trained network on its ability to predict good grasps for each of the test models, over a range of pose uncertainties. For each test model, the object was placed on the surface at a random position and orientation within the camera's field of view. The synthetically-rendered depth image, after applying the noise model, was then processed by the CNN to yield a grasp score for every pose, which was then convolved with the gripper's pose uncertainty to yield a robust grasp score for every pose. The gripper was then sent to the pose corresponding to the maximum over these scores, and a grasp was attempted. As with training, the grasp was scored as successful if the object remained within the gripper after it had moved to a height of 20cm from the surface.

We compared our method to two baselines. First, \emph{Centroids}, whereby the target gripper pose is that of the centroid of the image, with the gripper orientated perpendicular to the dominant direction, calculated via a PCA decomposition. Second, \emph{Best Grasp}, whereby the target pose is that of the maximum from our learned grasp function. Then, our method \emph{Robust Best Grasp}, corresponds to the maximum of the smoothed grasp function, which takes into account the pose uncertainty of the gripper by convolving the two functions. We experimented with varying levels of uncertainty in the gripper pose, by defining a diagonal covariance matrix, and setting the standard deviations in $(u, v)$ and $\theta$ to $\sigma_{uv}$ and $\sigma_\theta$ respectively. Based on this uncertainty and the target pose, a random pose was sampled for the true pose to be achieved by the gripper.

\begin{figure}[h]
\centering
 \includegraphics[width=\linewidth]{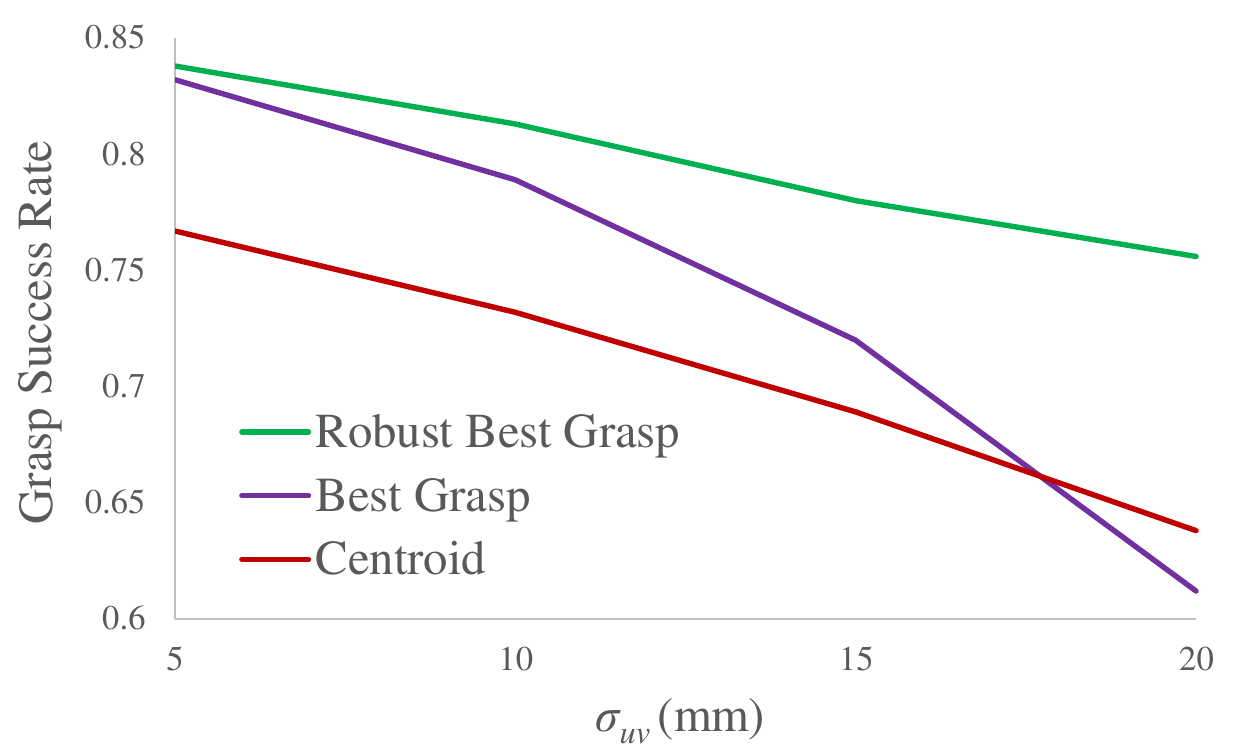}
 \caption{Plotting the data in Table \ref{table1} for an orientation uncertainty $\sigma_\theta$ of 10 degrees.}
 \label{fig:graph}
\end{figure}

Table \ref{table1} shows the results for all three methods over a range of pose uncertainty parameters. We see that as the uncertainty increases, both in terms of $(u, v)$ and $\theta$, the performance of each method degrades, as expected. Our method consistently outperforms the other methods, and the difference is particularly prominent at high levels of uncertainty. Figure \ref{fig:graph} shows the grasp success rate at a fixed orientation uncertainty of 10 degrees, whilst the uncertainty in $(u, v)$ is varied. We see that our \emph{Robust Best Grasp} method performs similarly to the \emph{Best Grasp} method at a low level of uncertainty, but as the uncertainty increases, \emph{Best Grasp} begins to result in poor grasps when only the maximum of the score function is taken without consideration of uncertainty. Interestingly, at very high levels of uncertainty, the \emph{Best Grasp} method actually performs worse than the \emph{Centroid} baseline, suggesting that when the gripper pose is highly unreliable, grasping the centroid of an object is a better strategy than attempting to pin point an optimum grasp.

\subsection{Real-World Grasping}

\begin{figure}
    \centering
    \includegraphics[width=0.7\linewidth]{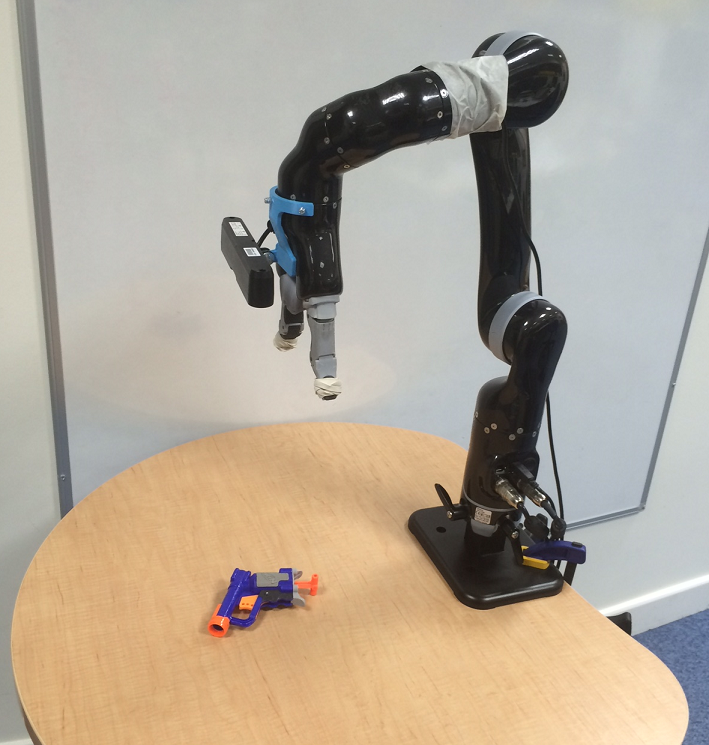}
    \caption{Our experimental setup for real-world grasping validation. The Kinova MICO arm has a Carmine 1.09 depth camera rigidly mounted to its wrist. The objects are then placed beneath the arm, within the field-of-view of the camera.}
    \label{fig:mico}
\end{figure}

\begin{figure}
\centering
 \includegraphics[width=\linewidth]{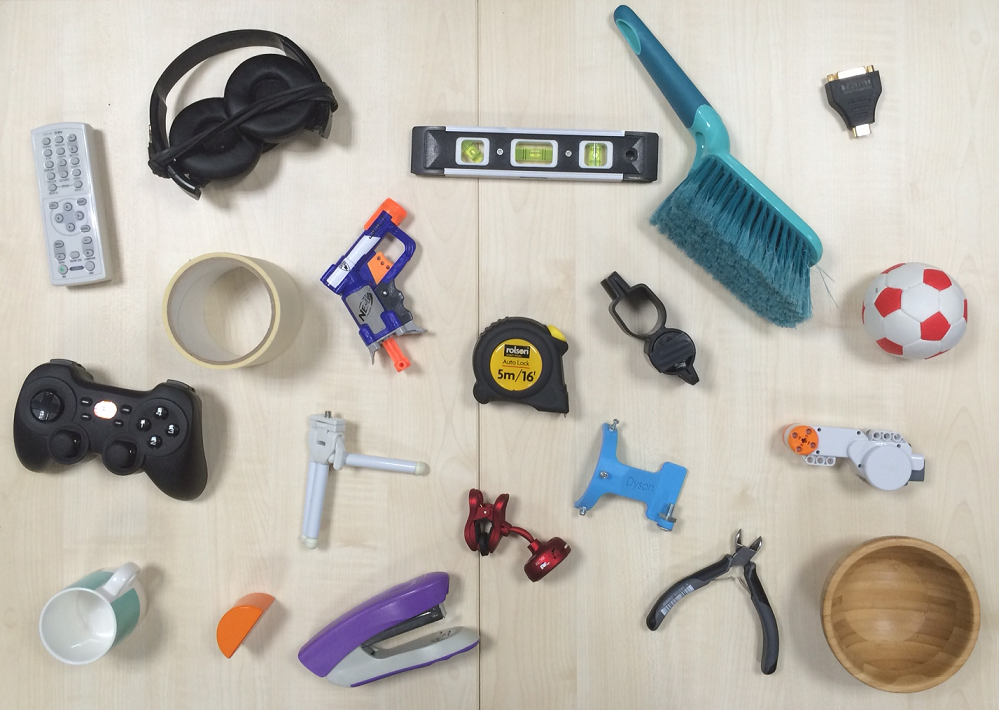}
 \caption{The set of 20 everyday objects used to evaluate our method on real-world grasping with a robot arm.}
  \label{fig:objects}
\end{figure}

To test how well our synthetic training data adapts to real-world robotics, both in terms of the synthetic depth images and the physics simulation, we conducted experiments on our Kinove MICO arm, as shown in Figure \ref{fig:mico}. We collated a set of 20 everyday objects, covering a broad range of sizes, shapes, and frictional coefficients, as displayed in Figure \ref{fig:objects}. Experiments were then conducted by placing each object at five random positions and orientations within a graspable area of the table upon which the robot was mounted. After processing the depth image with the network, kinematic controls were sent for the targeted grasp pose. As before, a grasp was counted as successful if it was lifted off the table to a height of 20cm.

We also experimented with varying the gripper pose uncertainty, First, we measured the true uncertainty of our arm, by sending commands to a variety of poses and recording the range of achieved poses. In practice, this could be achieved by a more thorough evaluation of the joint angle encoders to yield a more informative covariance matrix rather than simply a diagonal one. The uncertainty of our arm was measured to be $\sigma_{uv}$ = 6.2 pixels after transforming back into image space, and $\sigma_\theta$ = 4.7 degrees. Then, we increased the uncertainty to $\sigma_{uv}$ = 20 and $\sigma_\theta$ = 15 for further experiments.

Table \ref{table2} shows the results for the real-world experiments. We see that our robust method outperforms the two competing methods at the true pose uncertainty, and after increasing this uncertainty the relative performance of our method increases even further, as we saw during the synthetic experiments. Again, when pose uncertainty is very high, we see that the \emph{Centroid} method performs better than simply taking the maximum across the grasp function.

\begin{figure}
\centering
 \includegraphics[width=\linewidth]{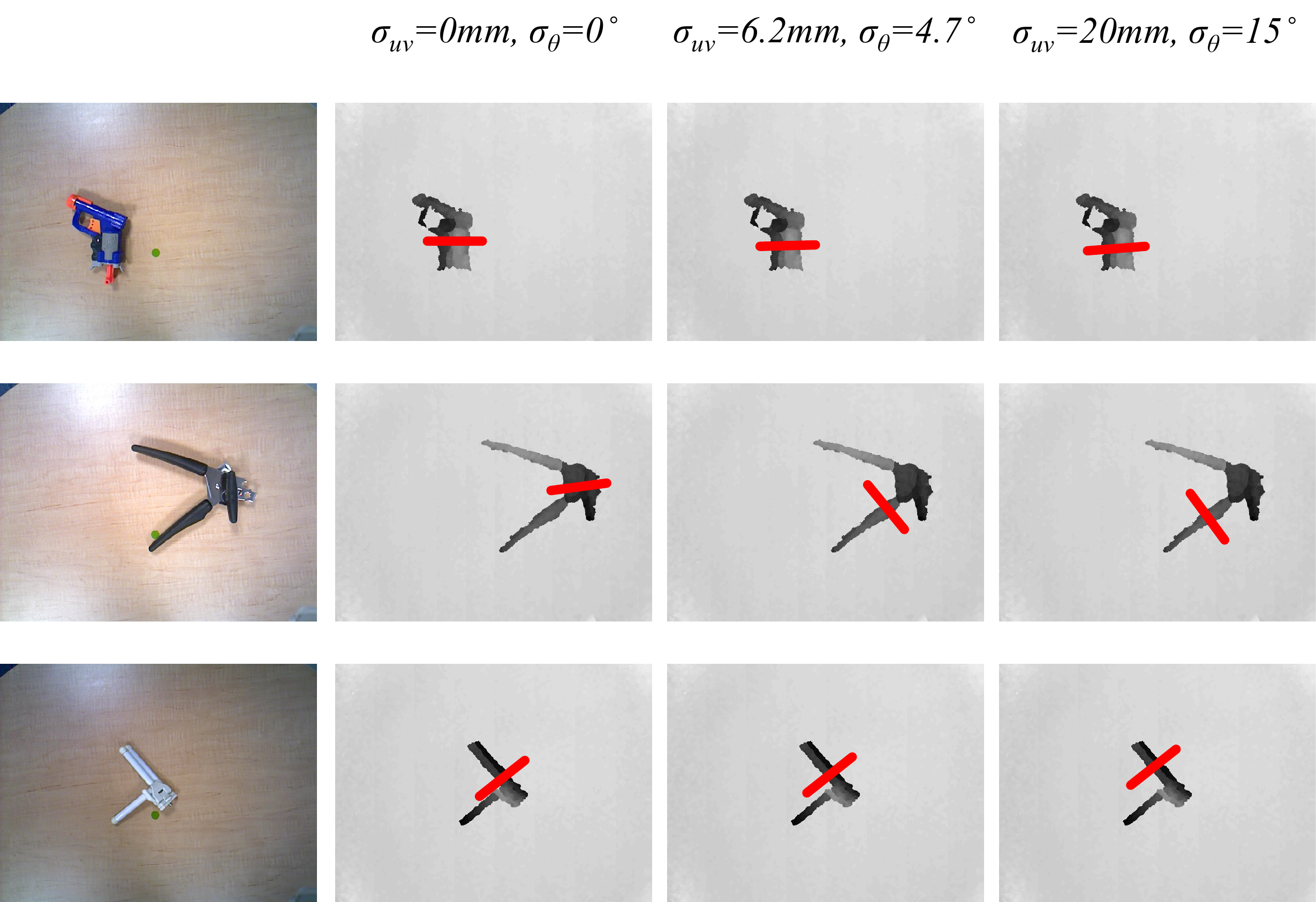}
 \caption{Demonstrating the effect of increasing gripper pose uncertainty. The image on the left shows an RGB observation of the object, with the three images towards the right showing the depth image. Each column represents a distinct gripper pose uncertainty which is assumed by the method, and used to smooth the predicted grasp function. The red line then shows the pose corresponding to the maximum of this smoothed grasp function, after performing trilinear interpolation.}
 \label{fig:example_grasps}
\end{figure}

\begin{figure}
    \centering
    \includegraphics[width=\linewidth]{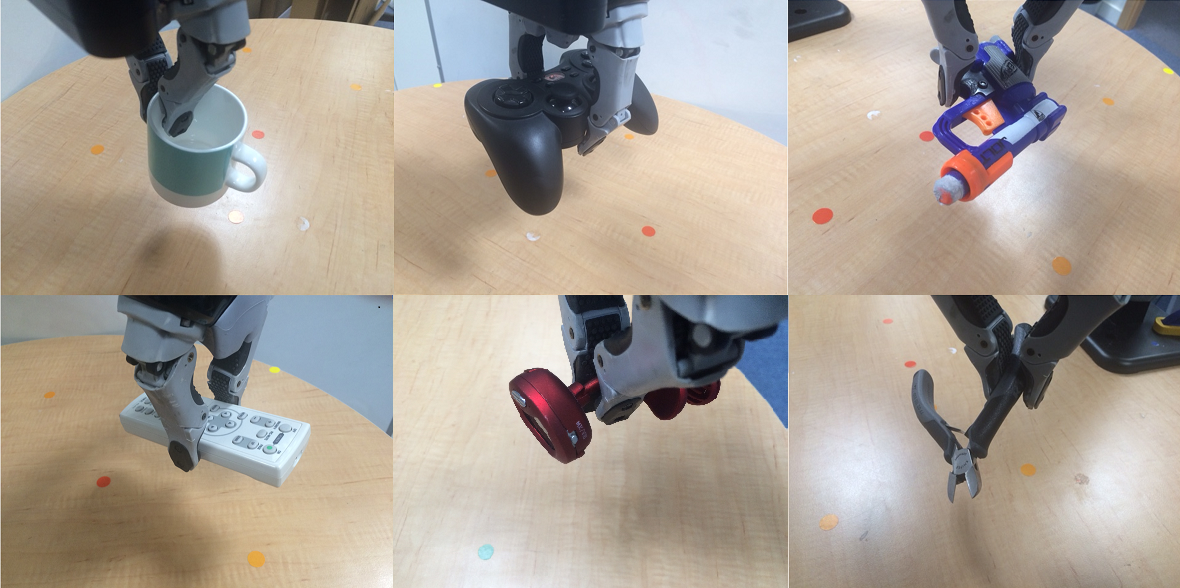}
    \caption{Examples of stable grasps achieved by our method.}
    \label{fig:real_grasps}
\end{figure}

Figure \ref{fig:example_grasps} then shows how increasing the uncertainty of the gripper pose affects the targeted grasp pose based on our robust method. With low uncertainty, the best pose is often at a stable grasp position as close to the centre of the object as possible. Then, as the uncertainty increases, the best pose tends to be repelled by regions of high object mass, to avoid any collisions with the object which might arise due to imprecise arm control. With very large uncertainty, the optimum grasp pose is often simply directly in the middle of a long, thin part of the object, regardless of whether this is close to the object's centre of mass. Finally, Figure \ref{fig:real_grasps} shows some example grasps executed by our method using the measured uncertainty of the robot.

\begin{table}
\normalsize
\centering
\begin{tabular}{c|cc}
\toprule
    Method & $\sigma_{uv}$=6.2, $\sigma_{\theta}$=4.7 & $\sigma_{uv}$=20, $\sigma_{\theta}$=15 \\
\midrule
Centroid & 0.752 & 0.648 \\
Best Grasp & 0.780 & 0.624 \\
Robust Best Grasp & 0.803 & 0.701 \\
\bottomrule
\end{tabular}
\caption{Real-world grasp success rates for the three implemented methods, for two different sets of pose uncertainties. Uncertainties $\sigma_{\theta}$ is in degrees, whilst $\sigma_{uv}$ is in mm.}
\label{table2}
\end{table}

\section{CONCLUSIONS}

In this work, we have developed a method for predicting a grasp quality score over all grasp poses, which we call the \emph{grasp function}. We investigated generating synthetic training data using physics simulation and depth image simulation, and using a CNN to map a depth image onto this grasp function. After convolving this grasp function with the gripper's pose uncertainty, we have shown that the pose corresponding to the maximum of this smoothed function is superior to the maximum of the original grasp function, both in synthetic and real-world experiments. 

The use of physics simulators and synthetic depth images has great capacity for extending this work to more complex tasks which also require large training datasets. For example, increasing the number of degrees of freedom in gripper pose, to incorporate height and angle-of-attack, would enable a greater range of grasps to be executed. However, with data generation already taking a week to complete, this would require a more refined selection of which simulations to process, and future work will investigate using active learning methods to enable the required scalability.

\section{ACKNOWLEDGEMENTS}

Research presented in this paper has been supported by Dyson Technology Ltd.

\end{document}